# Shapley variable importance cloud for machine learning models

Yilin Ning[1], Mingxuan Liu[1], Nan Liu[1,2,3,4*]

[1] Centre for Quantitative Medicine, Duke-NUS Medical School, Singapore

[2] Programme in Health Services and Systems Research, Duke-NUS Medical School, Singapore

[3] SingHealth AI Office, Singapore Health Services, Singapore

[4] Institute of Data Science, National University of Singapore, Singapore

*Correspondence: Nan Liu, Centre for Quantitative Medicine and Programme in Health Services and Systems Research, Duke-NUS Medical School, 8 College Road, Singapore, 169857. Phone: +65 6601 6503. Email: liu.nan@duke-nus.edu.sg

## Abstract

Current practice in interpretable machine learning often focuses on explaining the final model trained from data, e.g., by using the Shapley additive explanations (SHAP) method. The recently developed Shapley variable importance cloud (ShapleyVIC) extends the current practice to a group of "nearly optimal models" to provide comprehensive and robust variable importance assessments, with estimated uncertainty intervals for a more complete understanding of variable contributions to predictions. ShapleyVIC was initially developed for applications with traditional regression models, and the benefits of ShapleyVIC inference have been demonstrated in real-life prediction tasks using the logistic regression model. However, as a model-agnostic approach, ShapleyVIC application is not limited to such scenarios. In this work, we extend ShapleyVIC implementation for machine learning models to enable wider applications, and propose it as a useful complement to the current SHAP analysis to enable more trustworthy applications of these black-box models.

# Introduction

Interpretability is key to the implementation of prediction models in high-stakes decision making. Current practice in developing interpretable machine learning (ML) models, be it intrinsically interpretable models (e.g., regression or scoring models) or black box models with post hoc explanations (e.g., by using the Shapley additive explanations (SHAP) method (1)), centers around single models that optimize performance. The recently proposed variable importance cloud (VIC) method (2) was the first to extend the assessment beyond single models for more robust and less biased inference, which was demonstrated for logistic regression models, decision trees and convolutional neural network (via a surrogate logistic regression). Specifically, VIC investigates a group of models that have nearly optimal performance to understand the variability in variable importance and overall importance to the model class (instead of the single optimal model). Following from the VIC framework, we recently extended the well-received Shapley-based variable importance assessments to the Shapley variable importance cloud (ShapleyVIC) (3), which explicitly quantifies the uncertainty in overall variable importance for rigorous inference.

In view of the recent emphasis on intrinsic interpretability for high-stakes decision making (4,5), existing implementation of ShapleyVIC focused on transparent regression models, and the benefits from its robust variable importance assessments have been demonstrated for applications using logistic regression models (3) and regression-based scoring models (6). When working with tabular and static clinical data, such simple models are not necessarily outperformed by the complex machine learning models (7–9). However, ML (including deep learning) models have demonstrated preferable properties when working with data with more complex structures, e.g., longitudinal data (10), image input (11) or high-dimensional gene expression data (12,13). Commonly used ML models in healthcare research, e.g., support vector machine, extreme gradient boost, random forest and neural networks, are complex black boxes that heavily rely on post-hoc explanations for clinical interpretations (14,15). To facilitate trustworthy application of such powerful models, in this work we extend the application of ShapleyVIC to black-box ML models for more robust interpretations and hence more reliable clinical applications.

A key to the extension of the ShapleyVIC framework is a reasonable definition of "nearly optimal models" for different model classes. When working with regression models, ShapleyVIC defines nearly optimal models based on the multivariable normal distribution of regression coefficients that is estimated when training the optimal model, and all models in this multivariable normal space with model loss not exceeding a pre-defined threshold (e.g., 1.05 times the minimal loss) are viable candidates (3). While such definition builds on existing methodological works (2,16), the extension to ML models is non-trivial: general ML models no longer assess statistical distribution of parameters, and for neurol networks there are often too many relevant parameters to reasonably approximate the parameter space. In this work, we devise generalized definitions of nearly optimal models for broader classes of ML models that are relevant to model development practice, define ShapleyVIC values for these models for robust measures of overall variable importance, and demonstrate its benefit as a natural extension to the state-of-the-art SHAP analysis in real-data experiments.

## Methods

### ShapleyVIC overview

We first summarize the general ShapleyVIC framework, which is model agnostic (3). While conventional prediction tasks stop at training an optimal model by minimizing loss, ShapleyVIC follows the VIC framework (2) to analyze nearly optimal models from a Rashomon set, where model loss only exceeds the minimum loss by a small fraction (e.g., up to 5%) (2,16). Specifically, let $f(X; \theta)$ denote a prediction model for $Y$ using variables $X$ and parameters $\boldsymbol{\theta}$, and let $E\{L(f(X;\boldsymbol{\theta}), Y)\}$ denote the corresponding expected loss (e.g., cross entropy loss or mean squared error [MSE]). Within the model family, $f \in F$, there are infinitely many instances with different values for $\boldsymbol{\theta}$ (e.g., different coefficients for logistic regression models), and researchers are often interested in training an optimal model that minimizes expected loss, i.e.:

$$f^*(X; \boldsymbol{\theta}^*) = argmin_{f \in F} E\{L(f(X; \boldsymbol{\theta}), Y)\}.$$

Based on the optimal model, the set of nearly optimal models can be defined as (2,3,16):

$$R(\varepsilon, f^*, F) = \{f \in F | E\{L(f(X; \boldsymbol{\theta}), Y)\} \leq (1 + \varepsilon) E\{L(f^*(X; \boldsymbol{\theta}^*), Y)\}\},$$

where $\varepsilon = 5\%$ is an acceptable value.

As elaborated in previous works (2,3,5,16), analytically studying variable importance for all nearly optimal models, $R(\varepsilon, f^*, F)$, is challenging, and generating a representative sample of models from this set for empirical investigation is non-trivial. Instead, we empirically investigate a sample of nearly optimal models from $R(\varepsilon, f^*, F)$ that adequately represent the range of model loss of interest. For each nearly optimal model sampled from $R(\varepsilon, f^*, F)$, we quantify the global importance of each variable and the variability using

Shapley-based approaches, and use a random-effects meta-analysis approach to aggregate the information across all models, where the weighted average from the meta-analysis estimates overall variable importance, and the 95% prediction interval (PI) quantifies the uncertainty (3).

ShapleyVIC facilitates model interpretation using two forms of visual inferences, which we will demonstrate in an experiment in a later section. Similar to the state-of-the-art SHAP analysis (1) of the optimal model, which visualizes average variable impact on the model using bars and detailed local explanations using dots, ShapleyVIC summarizes overall variable importance in a bar plot with statistical significance indicated by error bars, and highlights the uncertainty in variable importance across models using a coloured violin plot to facilitate further explorations (3).

**ShapleyVIC for machine learning (ML) models**

In the initial development (3), detailed implementation and application has been described for regression models, where nearly optimal models are generated using a rejection sampling approach based on the multivariable normal distribution of regression coefficients, and variable importance to each sampled model is quantified using Shapley values, specifically the Shapley additive global importance (SAGE) approach (17) that has built-in variability measure. In this work, we extend ShapleyVIC application for general ML models to enable wider applications, where predictions are not restricted to simple linear functions of predictors. We use SHAP instead of SAGE to quantify variable importance from nearly optimal ML models, which better aligns with the current practice in IML, and provides opportunities for seamless extensions beyond tabular data (e.g., to image classifications (18)). In the following subsections, we describe our proposed methods to generate nearly optimal ML models, using an application to multiplayer perceptron (MLP) as an example.

*ShapleyVIC values for MLP*

The nearly optimal set $R(\varepsilon, f^*, F)$ for MLP is composed in two stages, with both endogenous and exogenous perturbation towards the optimal model. Instead of directly obtaining $f^*$, we add an additional $L_2$ penalty term to the loss (endogenous perturbation), shifting the objective function with the control of coefficient $\lambda$ (19). For each $\lambda \geq 0$, we can obtain corresponding $f'(X; \boldsymbol{\theta}', \lambda)$, as pivots to explore the neighborhood region of $f^*$:

$$f'(X; \boldsymbol{\theta}', \lambda) = argmin_{f \in F} E\{L(f(X; \boldsymbol{\theta}), Y) + \lambda L_2(f)\}.$$

Note that $f'(X; \boldsymbol{\theta}', \lambda)$ is equal to $f^*(X; \boldsymbol{\theta}^*)$ when $\lambda = 0$. As a constraint, the loss threshold limits the diversity of the subset $F' = \{f'(X; \boldsymbol{\theta}', \lambda) | \lambda \geq 0\}$ in case it reaches the boundary where the models differ notably from $f^*$ in terms of model loss. The nearly optimal models $R_1(\varepsilon, f^*, F)$ obtained at this stage is:

$$R_1(\varepsilon, f^*, F) = \{f' \in F' | E\{L(f'(X; \boldsymbol{\theta}', \lambda), Y) + \lambda L_2(f')\} \leq (1 + \varepsilon) E\{L(f^*(X; \boldsymbol{\theta}^*), Y)\}\},$$

and models randomly generated from this set will serve as "seeds" for the expanded search in the next stage.

At the second stage, we expand samples from $R_1(\varepsilon, f^*, F)$ into a larger set of models by fine-tuning based on subsets of the training set. Fine-tuning of pretrained neural networks on additional data is a commonly used technique in image analysis, with the aim to transfer existing models to new applications instead of training new models from scratch (20–23). In this work, we use subsets of the training set with intentionally diversified sample characteristics as a means of exogenous perturbation, i.e., to guide the expansion of "seed" models in $R_1$ to explore a wider range in the space of "nearly optimal models". We construct the subsets variable-wise, i.e., by splitting the training set into subsets defined by one variable each time (where continuous variables are first categorized based on quantiles). After fine-tuning "seed" models based on these various subsets with limited epochs, we harvest a larger set of models that explore a wider range of the model space around the optimal model, $f^*$, and reject models with loss exceeding the pre-specified threshold to obtain a final sample of models from the original model space of interest, i.e., $R(\varepsilon, f^*, F)$.

As motivated earlier, we derive overall variable importance measures for MLP based on SHAP values from sampled nearly optimal models. Following from IML convention, we use the mean absolute SHAP values for a MLP, evaluated on an explanation data, as global variable importance measures, and quantify their uncertainty using the standard errors of absolute SHAP values across the explanation data (i.e., the standard deviation of absolute SHAP values divided by the square root of the sample size of the explanation data). The ShapleyVIC values as overall importance measures are derived by applying a random-effect meta-analysis to such importance measures from all sampled nearly optimal models, and the same bar plot and violin plot used in ShapleyVIC analysis of regression models can be used to visualize findings for MLP.

**ShapleyVIC-based ensemble ranking**

In the initial ShapleyVIC development with regression models and an application with logistic regression predictions (6), we described an ensemble variables ranking approach that accounts for the variability in ShapleyVIC values. In this work, we formally describe the ShapleyVIC-based ensemble ranking for general ML models.

Consider ShapleyVIC analysis of a prediction task using ML model class $F$ (e.g., the class of MLP) and $d$ candidate variables, and let $f^* \in F$ denote the optimal model trained from all $d$ variables. For any nearly optimal model $f \in R(\varepsilon, f^*, F)$ generated using the algorithm described in the previous section, let $\widehat{mr}_j^s(f)$ denote the ShapleyVIC value of the $j$-th variable ($j = 1, \ldots, d$) for this model, and let $\hat{\sigma}_j(f)$ denote the estimated variability associated with $\widehat{mr}_j^s(f)$. As defined in the previous subsection, for MLP $\widehat{mr}_j^s(f)$ is the mean absolute SHAP values for the $j$-th variable, and $\hat{\sigma}_j(f)$ is the standard error.

Assuming independent normal distributions for ShapleyVIC values between any pair of variables for model $f$, for each variable, $X_j$, we assess if it is significantly more important than variable $X_k$ ($\forall k \neq j$) based on the normal distribution of the difference in ShapleyVIC values:

$$\Delta_{jk} = \widehat{mr}_j^s(f) - \widehat{mr}_k^s(f) \sim N\left(mr_j^s(f) - mr_k^s(f), \sigma_j^2(f) + \sigma_k^2(f)\right).$$

$X_j$ is significantly more important than $X_k$ for model $f$ if and only if the difference is positive and significantly different from zero, i.e.:

$$\Delta_{jk} > 0 \text{ AND } \left|\frac{\Delta_{jk}}{\sqrt{\sigma_j^2(f) + \sigma_k^2(f)}}\right| > 1.96.$$

We compare all possible pairs (in total $\binom{d}{2} = (d(d-1))/2$) of variables, and subsequently rank the $d$ variables to model $f$ based on the number of times each variable is significantly more important than the other $d - 1$ variables. Assigning rank 1 to the most important variable and using the same smallest integer value available for tied ranks, we rank the $d$ variables for each model, and use the average rank of each variable across all nearly optimal models to generate an ensemble ranking for all candidate variables.

**Experiments**

To demonstrate our proposed extension of ShapleyVIC for ML models, we apply it with multilayer perception (MLP) in a study that predicts 30-day mortality using 17 variables (including patient demographics, inpatient admission type, laboratory tests, vital sign, and historical healthcare utilization) extracted from the electronic health records of 46,318 patients who visited the emergency department of the Singapore general hospital in 2017 and where subsequently admitted to the hospital. This study was approved by Singapore Health Services' Centralized Institutional Review Board (CIRB 2021/2122), and a waiver of consent was granted for electronic health record data collection. We assess the overall variable importance using ShapleyVIC, and compare it with a conventional SHAP analysis of the optimal model in shortlisting the most important variables using the parsimony plot. As explained in our previous work (3,6), at most a few thousand samples are needed to generate ShapleyVIC values, therefore in our experiment we first generated a random sample of approximately 2500 subjects as explanation data, and then randomly split the remaining samples into training (80%) and validation (20%) sets.

**Results**

The training, validation and explanation data consisted of 35,066, 8766 and 2486 subjects, respectively. Figure 1 visualizes overall variable importance from the ShapleyVIC analysis of MLP (Figure 1(A)), and the variability of variable importance across models (Figure 1(B)). Age, blood pressure and creatine were found to be more important than other variables in predicting 30-day mortality, and number of ED visits in the past year had higher overall importance than other historical healthcare utilization in the past year, i.e., the number of surgeries, ICU stays, or high dependency ward (HD) stays. Gender and race had overall importance close to zero. While these findings are similar to those from the SHAP analysis of the optimal MLP (see Figure 2), the additional uncertainty intervals (specifically, the 95% PIs) of overall variable importance provide additional insights that help nurture trust in ML models, e.g., to direct attention to variables with significantly different overall importance and justify slight perturbations among variables with overlapping intervals.

While the variables with high overall importance based on ShapleyVIC (see Figure 1(A)) also tended to have high global importance based on SHAP analysis of the optimal MLP (see Figure 2), findings from these two approaches differ slightly in variable ranking, specifically based on mean absolute SHAP values and ShapleyVIC-based ensemble ranking, as visualized in Figure 3. These differences in ranking resulted in a smoother parsimony plot based on the ShapleyVIC analysis compared to that based on SHAP analysis (see Figure 4), especially for the top few variables, suggesting that the ShapleyVIC-based variable ranking may better reflect the contribution of variables to predictions compared to a SHAP analysis of just the optimal model.

**Figure 1.** ShapleyVIC analysis of MLP in the 30-day mortality study.

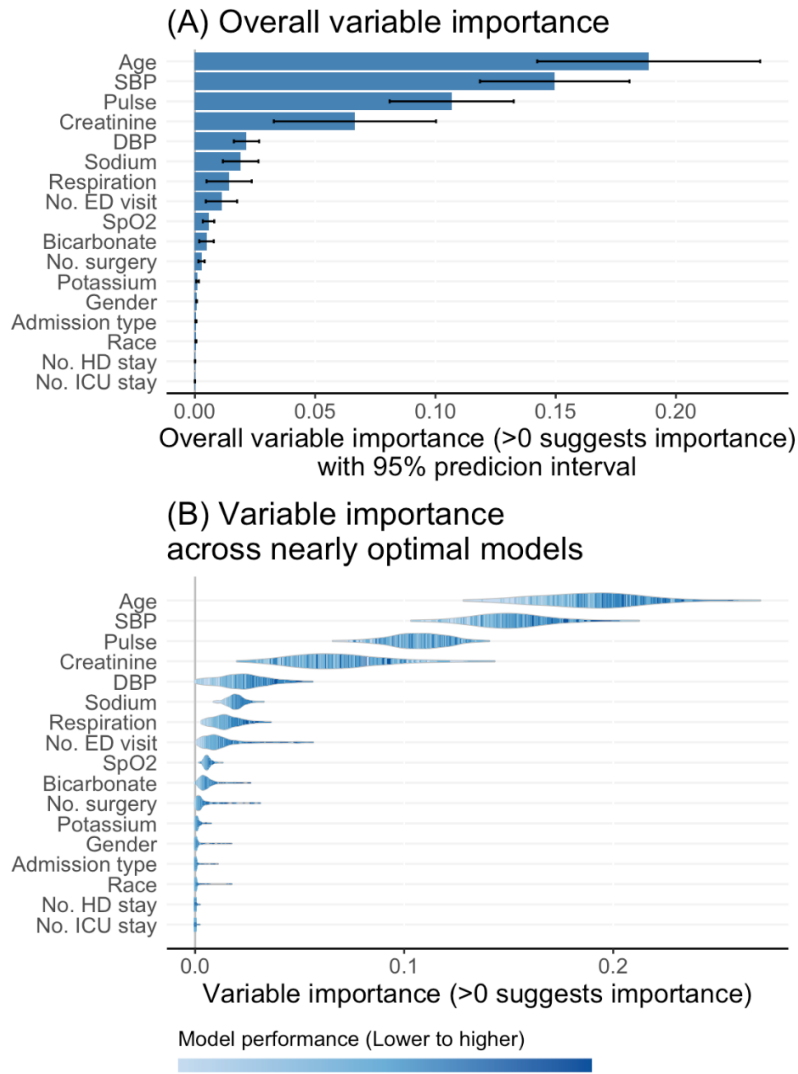

**Figure 2.** SHAP analysis of global variable importance (measured by mean absolute SHAP values) to the optimal MLP in the 30-day mortality study.

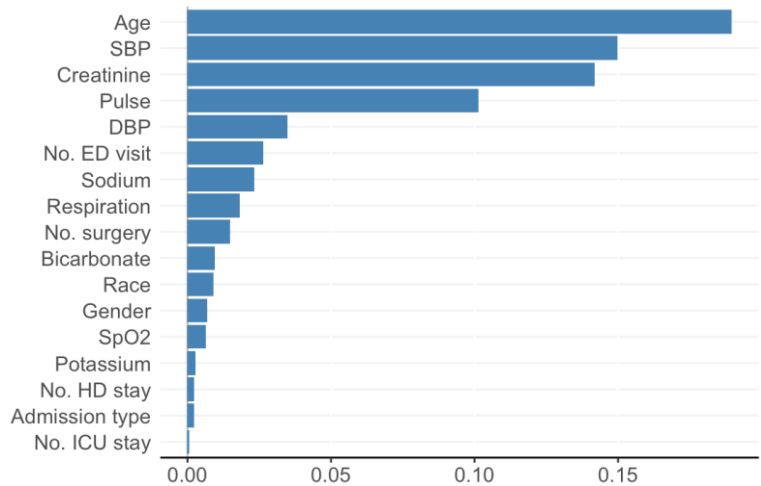

**Figure 3.** Variable ranking in the 30-day mortality study by mean absolute SHAP values and ShapleyVIC-based ensemble ranking.

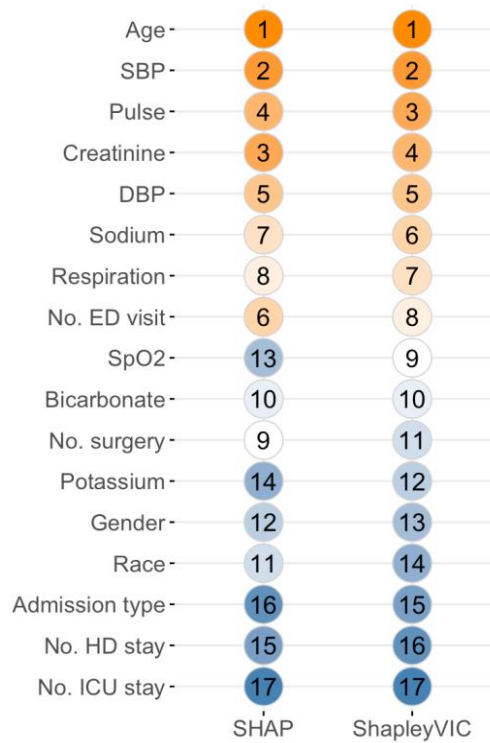

**Figure 4.** Parsimony plot for MLP of mortality in the 30-day mortality study based on variable ranking from SHAP and ShapleyVIC.

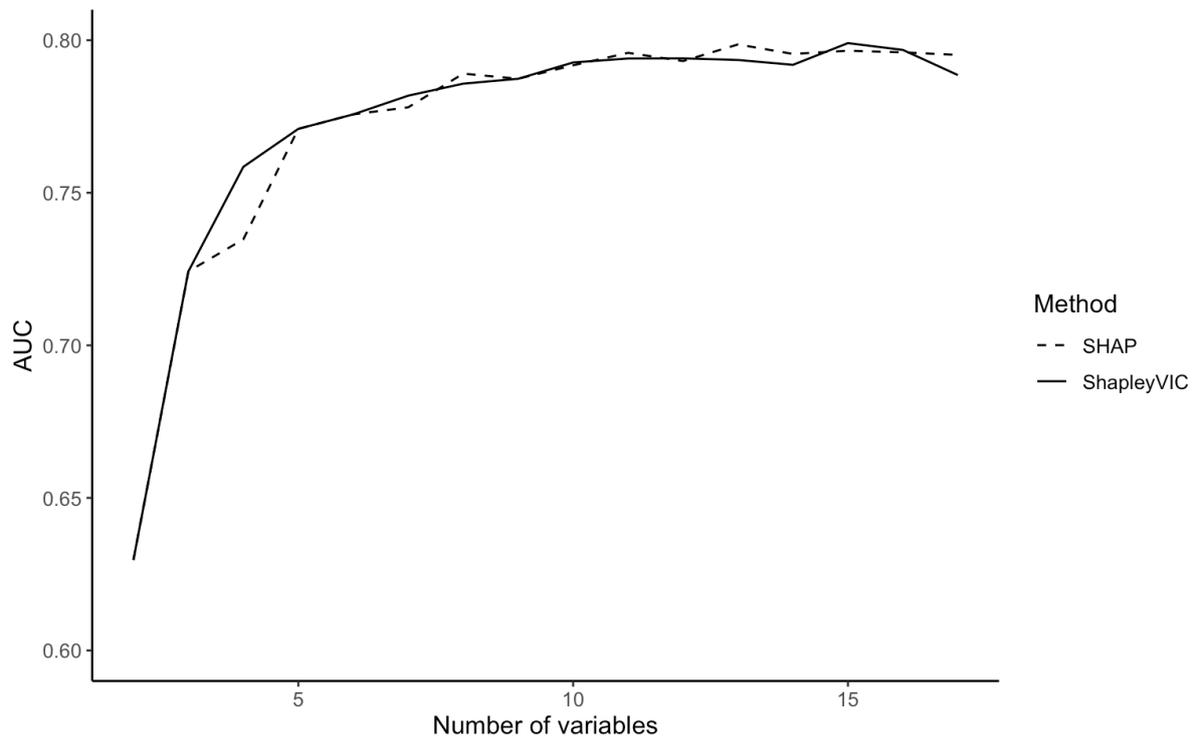

## Discussion

Understanding the uncertainties involved in data-driven analyses is important to appropriate interpretation of the findings. Current practice in explaining ML prediction models is dominated by post hoc model evaluation of variable importance without uncertainty intervals, e.g., by using SHAP values aggregated across the data. Our recent development of ShapleyVIC extends the well-received Shapley values to take into account the uncertainties in variable contributions to accurate prediction, generating robust overall variable importance measures to help nurture trust in the findings. This work further extends ShapleyVIC to accommodate general ML models for broader applications.

Our initial development demonstrated the feasibility and usefulness of the ShapleyVIC framework in applications for logistic regression analyses, using the SAGE method to quantify variable importance for each nearly optimal logistic regression model. In this work, we have broadened the application of the ShapleyVIC workflow to explain general ML models with complex non-linear mathematical structures and used SHAP as the variable importance measure, and described our strategy for generating and analyzing nearly optimal MLPs as an example. In our real-data experiment, we presented the extended ShapleyVIC analysis as a useful complement to the well-received SHAP analysis, where the uncertainty intervals of variable importance reported by ShapleyVIC can help domain experts differentiate random noise from noteworthy findings, hence better appreciate and eventually develop trust in ML models. Moreover, by aggregating information across models (instead of basing on the single optimal model), the overall variable importance estimated using the extended ShapleyVIC method somewhat better reflected variable contributions to prediction than the conventional SHAP analysis of the optimal model, especially for the ones with higher importance. Hence, ML applications may benefit from additional ShapleyVIC analysis in complementary to a conventional SHAP explanation of the final model for more in-depth understanding of variable contributions to predictions.

The use of SHAP (instead of SAGE) to compute variable importance for each nearly optimal model enables smoother integration with current practice in healthcare applications. A disadvantage of ShapleyVIC compared to SHAP is the longer run time due to the need to investigate a group of nearly optimal models for richer and more robust inference. We are actively exploring for modifications to smoothen ShapleyVIC implementation. For example, we are investigating whether it is feasible to provide reliable overall importance and uncertainty measures from less than 350 nearly optimal models and smaller explanation sets to reduce run time. The current definition of "nearly optimal" is based on model loss that may not be straightforward when communicated to domain experts, who may be more familiar with performance metrics such as the area under the receiver operating characteristic curve. In future work we will modify our definition of "nearly optimal" based on more intuitive metrics. We also seek to extend the ShapleyVIC methodology to a broader class of ML models, e.g., to general deep learning models to enable applications for image data.